\documentclass[runningheads]{llncs}
\usepackage{graphicx}
\usepackage{graphicx}
\usepackage{amsmath}
\usepackage{amssymb}
\usepackage{booktabs}
\usepackage{amsmath}
\usepackage{amssymb}
\usepackage{bm}
\usepackage{subfigure}
\usepackage{booktabs}
\usepackage{multirow}
\usepackage{booktabs}
\usepackage{caption}
\usepackage{float}
\usepackage{algorithmic}
\usepackage{algorithm}
\usepackage{bbding}
\usepackage{array}
\usepackage{color}
\usepackage{diagbox}
\usepackage{threeparttable}
\usepackage{pifont}
\usepackage{tikz}
\usepackage{amsmath,amssymb} 
\usepackage{color}
\usepackage{amssymb}%
\usepackage{comment}
\usepackage[accsupp]{axessibility}  %
\usepackage{bbding}

\begin{document}

\title{TAPE: Task-Agnostic Prior Embedding for Image Restoration} 

\titlerunning{TAPE: Task-Agnostic Prior Embedding for Image Restoration}

\author{Lin Liu\textsuperscript{1} \hspace{3mm}
    Lingxi Xie\textsuperscript{3} \hspace{3mm}
    Xiaopeng Zhang\textsuperscript{3} \hspace{3mm}
   Shanxin Yuan\textsuperscript{4} \\ 
Xiangyu Chen\textsuperscript{5,6} \hspace{3mm}
Wengang Zhou\textsuperscript{1,2} \hspace{3mm}
Houqiang Li\textsuperscript{1,2} \hspace{3mm}
Qi Tian\textsuperscript{3}}
\authorrunning{L. Liu, L. Xie, X. Zhang, S. Yuan, X. Chen, W. Zhou, H. Li, Q. Tian}
\institute{\footnotesize{$^1$CAS Key Laboratory of Technology in GIPAS,\\ EEIS Department,
University of Science and Technology of China} \\ \footnotesize{$^2$Institute of Artificial Intelligence, Hefei Comprehensive National Science Center} \\ \footnotesize{$^3$Huawei Cloud BU \qquad  $^4$Huawei Noah's Ark Lab \qquad  $^5$University of Macau} \\
\footnotesize{$^6$Shenzhen Institutes of Advanced Technology, CAS}}

\maketitle

\begin{abstract}
Learning a generalized prior for natural image restoration is an important yet challenging task. Early methods mostly involved handcrafted priors including normalized sparsity, $\ell_0$ gradients, dark channel priors, etc. Recently, deep neural networks have been used to learn various image priors but do not guarantee to generalize. 
In this paper, we propose a novel approach that embeds a task-agnostic prior into a transformer.
Our approach, named Task-Agnostic Prior Embedding (TAPE), consists of two stages, namely, task-agnostic pre-training and task-specific fine-tuning, where the first stage embeds prior knowledge about natural images into the transformer and the second stage extracts the knowledge to assist downstream image restoration. Experiments on various types of degradation validate the effectiveness of TAPE. The image restoration performance in terms of PSNR is improved by as much as 1.45dB and even outperforms task-specific algorithms. 
More importantly, TAPE shows the ability of disentangling generalized image priors from degraded images, which enjoys favorable transfer ability to unknown downstream tasks.
\end{abstract}

\section{Introduction}
A good image prior can help to distinguish many kinds of noises from original image contents and improve the quality of images.
Learning %
an image prior is important and challenging for image restoration tasks. %
Early studies explore specific degradation priors to achieve good performances on some low-level vision tasks, such as image dehazing~\cite{he2010single,zhu2015fast}, image deblurring~\cite{pan2016blind,levin2009understanding}, and image deraining~\cite{li2016rain,zhu2017joint}.
However, most priors are hand-crafted and mainly based on limited observations. 
With the popularity of deep learning, 
data-driven image priors estimated by combining conventional degradation properties with deep neural networks have been explored~\cite{li2019blind,guo2021joint,zhang2017learning,HunsangLee2020UnsupervisedLI,AlonaGolts2020UnsupervisedSI,Pan_2020_CVPR,Chen_2019_CVPR}.
But these networks capturing task-specific priors, do not guarantee to generalize to unseen tasks.
\begin{figure*}[!t]
		\centering
		\includegraphics[width=0.96\textwidth]{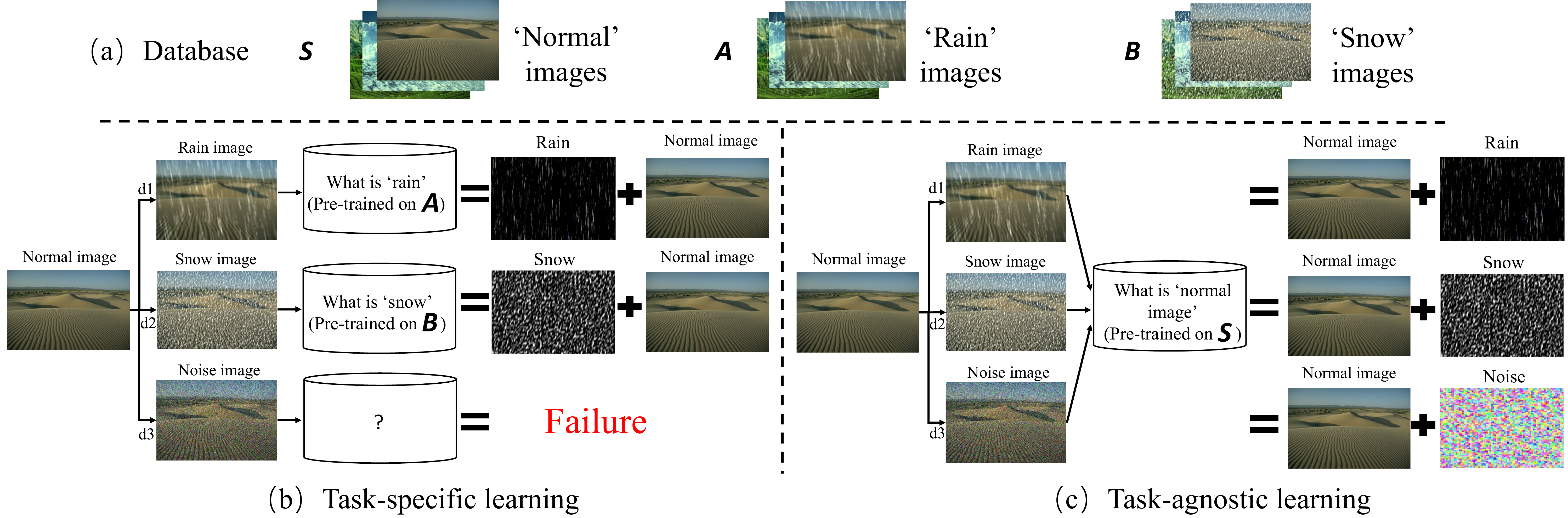}
		\caption{The illustration of the differences of task specific learning and task-agnostic learning. Our method aims to learn `what is normal image' instead of `what are rain, snow or other degradation'.}
		\label{illust}
\end{figure*}

Recently, there are also efforts in learning complicated priors for low-level vision tasks~\cite{ulyanov2018deep,gu2020image,pan2020exploiting,chen2020pre,li2020learning,yu2018crafting}. These methods can be roughly grouped into two types. The first type~\cite{chen2020pre,li2020learning} learns specific priors for each task, \textit{i.e.}, the priors formulate `what is the distribution of specific noise'. Despite their effectiveness, these methods are often difficult to transfer across different tasks. The second type instead formulates generalized image priors, \textit{i.e.}, `what is the distribution of normal images'. For this purpose, these methods~\cite{gu2020image,pan2020exploiting,Wang_2021_CVPR} often make use of scalable GANs~\cite{AndrewBrock2018LargeSG} pre-trained on natural images, hence, the learned priors are often hidden in a latent feature space, making it difficult to disentangle the noise from image contents, especially in the scenarios of complex images.

In this paper, we propose Task-Agnostic Prior Embedding (TAPE), a novel kind of priors that are easily generalized to different low-level vision tasks. An intuitive comparison between TAPE and prior task-specific learning is illustrated in Fig.~\ref{illust}. TAPE absorbs the benefits of the aforementioned approaches: on the one hand, we learn the distribution of normal images from non-degraded natural images, which does not rely on any true or synthesized degradation; on the other hand, the priors are encoded in a simple prior learning module named PLM and the main network can decode them by transformer decoders (query embeddings). 
The training procedure of TAPE consists of two stages, namely, task-agnostic pre-training and task-specific fine-tuning, where a pixel-wise contrastive loss is designed in the first stage for unsupervised low-level representation learning.

In the experiment, we pre-train our model on four tasks (including deraining, deraindrop, denoising, and demoireing), fine-tune and test it on these four known tasks and four unknown tasks (desnowing, shadow removal, super-resolution, and deblurring).
After the one-time learning, the generalized image prior (through pre-training) can be transferred to different tasks.
Quantitative and qualitative experimental comparisons show that the proposed TAPE improves the performance for multiple tasks in both task-specific and task-agnostic settings. 
In particular, our method improves the PSNR by 1.45dB, 1.03dB, 0.84dB, 0.49dB, and 0.75dB on the
Rain200L, Rain200H, Raindrop800, SIDD, and TIP2018 datasets, respectively.
The task-agnostic pre-training without touching the real noisy image on SIDD increases PSNR by 0.31dB.
For the unseen tasks in the pre-training, TAPE improves the PSNR by 0.91dB, 0.29dB, 0.41dB and 0.48dB on desnowing, shadow removal, super-resolution, and deblurring, respectively.

In summary, the contributions of our work are: 
\begin{itemize}
  \item The possibility and importance of learning task-agnostic and generalized image prior is addressed. As far as we know, TAPE is the first work to (explicitly) represent the universal prior that can be used in multiple image restoration tasks. We disentangle the generalized clean image prior of the corrupted images from the degrading objects/noises.
  \item We propose a two-stage method named TAPE for image restoration to learn the generalized degradation prior. The experiments demonstrate that our method can be easily applied to pre-train image restoration algorithms with other transformer backbones.
  \item We propose a pixel-wise contrastive loss in pre-training for learning better generalized features for PLM, which increases the generalization ability.
\end{itemize}

\section{Related Work}

\textbf{Image Restoration.} Image restoration is a general term for a series of low-level vision tasks, including denoising~\cite{zhang2017beyond,zhang2018ffdnet,guo2019toward}, deraining~\cite{li2016rain,zhu2017joint,li2018recurrent}, deblurring~\cite{li2019blind,kupyn2019deblurgan}, demoireing~\cite{sun2018moire,zheng2020image,he2019mop}, \textit{etc}. 
The aim of image restoration is to restore clean $\mathbf{x}$ from corrupted $\mathbf{y}$. 
The corrupted image $\mathbf{y}$ can be formulated as, 
$\mathbf{y}=\mathbf{H} \mathbf{x}+\mathbf{v}$, where $\mathbf{H}$, $\mathbf{x}$, and $\mathbf{v}$ are degradation matrix, underlying clean image, and noise, respectively.
Before the deep learning era, studies design hand-crafted features of the degradation objects (\textit{e.g.}, rain, snow, \textit{etc.}) for different image restoration tasks. 
With the popularity of convolutional neural networks (CNNs), a handful of deep-learning based methods are proposed to handle one or multiple types of image restoration tasks. Most of these methods design task-specific models or loss functions to achieve better performances. %
For image super-resolution \cite{dong2015image,isobe2020video}, Dong \textit{et al.} propose SRCNN~\cite{dong2015image} to obtain high-resolution images from the corresponding low-resolution images. 
For HDR imaging, solutions \cite{domainplus,dai2021wavelet,li2022sj} are proposed for using multiple exposed images to reconstruct an HDR image. Fu \textit{et al.}~\cite{fu2017removing} introduce a ResNet-based CNN for image deraining. Li~\textit{et al.} and Yu~\textit{et al.}, propose FDRNet~\cite{li2020learning} and RL-Restore~\cite{yu2018crafting} to handle hybrid-distorted image restoration tasks. Zheng \textit{et al.}~\cite{zheng2021learning} propose a learnable bandpass filter network for image demoireing. 
Different from these methods, we explore the power of pre-training to handle several image processing tasks.

\textbf{Image Degradation Prior and Natural Image Prior.}
Since image restoration is ill-posed, the image prior can help to constrain the solution space. From the Bayesian perspective, the solution $\hat{x}$ can be obtained by optimizing:
\begin{equation}
\hat{\mathbf{x}}=\arg \min _{\mathbf{x}} \frac{1}{2}\|\mathbf{y}-\mathbf{H} \mathbf{x}\|^{2}+\lambda \Phi(\mathbf{x}),
\label{eq:opt}
\end{equation}
where the first term is the fidelity and the second term is the regularization.
Deep-learning based methods try to learn the prior parameters $\Theta$ and a compact inference through an optimization of a loss function on a training dataset with corrupted-clean image pairs. Then Eqn.~\ref{eq:opt} can be refined as,
\begin{equation}
\min _{\Theta} \ell(\hat{\mathbf{x}}, \mathbf{x}) \hspace{0.8em} \text {s.t.}\hspace{0.8em} \hat{\mathbf{x}}=\arg \min _{\mathbf{x}} \frac{1}{2}\|\mathbf{y}-\mathbf{H} \mathbf{x}\|^{2}+\lambda \Phi(\mathbf{x} ; \Theta).
\end{equation}

Image priors have been widely used in computer vision, including markov random fields~\cite{roth2005fields,zhu1997prior}, dark channel prior~\cite{he2010single,pan2016blind}, low rank prior~\cite{chen2013generalized}, and total variation~\cite{rudin1992nonlinear,babacan2008variational}. 
He \textit{et al.}~\cite{he2010single} propose dark channel prior for image dehazing. It exploits the prior property that in an haze-free image there are pixels where at least one color channel is of low value. Chen \textit{et al.}~\cite{chen2013generalized} propose a low-rank model to capture the spatial and temporal correlations between rain streaks.
Different from these task-specific priors, we use a pre-trained network to extract more general priors from images. The prior queries are also being adjusted during end-to-end training.
Recently, deep image prior (DIP)~\cite{ulyanov2018deep} shows that image statistics can be implicitly captured by the network's structure, which is also a kind of prior. %
Inspired by DIP, some attempts use a pre-trained GAN as a source of image statistics~\cite{gu2020image,pan2020exploiting,richardson2020encoding,elhelou2020bigprior,chan2020glean,Wang_2021_CVPR}. MGAN prior~\cite{gu2020image} utilizes multiple latent codes to increase the power of the pre-trained GAN model. 
DGP~\cite{pan2020exploiting} fine-tunes the weights generator together with the latent code and use the discriminator to calculate the gap between the generated and real images.

\textbf{Image Restoration Transformers.}
Transformer~\cite{vaswani2017attention} is a new type of neural network framework of using mainly self-attention mechanism. 
It has achieved many successes in computer vision tasks, including object classification~\cite{dosovitskiy2020image}, objection detection~\cite{Zhu2021DeformableDD,Dai2021UPDETRUP,NicolasCarion2020EndtoEndOD}, \textit{etc}.
Recently, it's also been used in image restoration tasks~\cite{yang2020learning,chen2020pre,zeng2020learning,liang2021swinir,wang2021uformer,edtli,XiangyuChen2022ActivatingMP,Liu2022SiamTransZM}. 
Chen \textit{et al.}~\cite{chen2020pre} and Li \textit{et al.}~\cite{edtli} develop pre-trained transformers IPT and EDT respectively for some low-level vision tasks.
Wang \textit{et al.}~\cite{wang2021uformer} and  Zamir~\textit{et al.}~\cite{Zamir2021Restormer} design novel transformer structures, named Uformer and Restormer respectively, and obtain good performance on several image restoration tasks.
Most of these methods proposed new transformer-based backbones for image restoration, whereas the goal of our work is a new pre-training pipeline for image restoration.
One can apply our method to pre-train an image restoration algorithm with other backbones.
We make use of the strong fitting ability of transformers and the learning ability of the transformer decoder to embed prior information.

\section{Proposed Method}
In this section, we first illustrate why we need to learn a task-agnostic prior (Sec. 3.1).
And then, the architecture (Sec. 3.2) and the pipeline (Sec. 3.3) of our method are presented.
At last, we briefly discuss the relationship to prior work (Sec. 3.4).

\subsection{Motivation}

The motivation is shown in Fig.~\ref{illust}, where we assume that many image restoration tasks (\textit{e.g.}, deraining, desnowing, \textit{etc.}) are present. Most existing image restoration methods learn a specific model for each single task, trying to capture task-specific priors (\textit{e.g.}, what patterns are likely to be rain and thus need to be removed). While the learning procedure is straightforward, the trained models are difficult to be applied to new restoration tasks because the degradation patterns may have changed significantly.

To alleviate the burden, we propose a different pipeline that is generalized across different restoration tasks. The key is to learn \textbf{task-agnostic priors} (\textit{i.e.}, what patterns are likely to be from clean, non-degraded images) rather than the aforementioned, task-specific counterparts. For this purpose, we embed an explicit module, the prior learning module (PLM), into the network, and design a two-stage learning procedure that (i) pre-trains the architecture on multi-source clean images and then (ii) fine-tunes it on specific image restoration datasets. The second stage often occupies a small portion of computation, showing the advantage of our method.

\begin{figure*}[!t]
		\centering
		\includegraphics[width=0.88\textwidth]{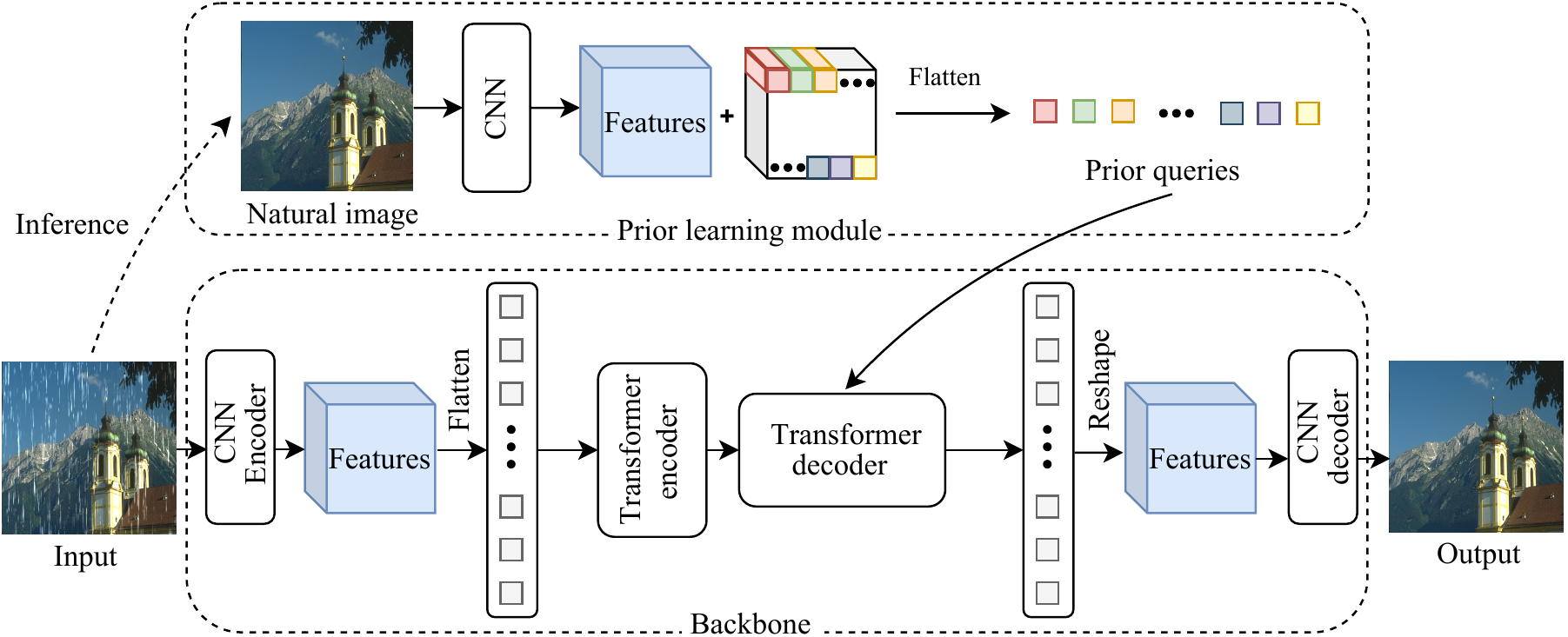}
		\caption{The network architecture of our TAPE-Net. It consist of two parts: Backbone and prior learning module. With the input of natural images, PLM learns the features that natural images contain, not the features that noise contains. This makes our approach task-agnostic.}
		\label{network_ipt}
\end{figure*}

\subsection{Network Architecture}

The TAPE-Net (see Fig. \ref{network_ipt}) consists of two components: backbone and prior learning module. The backbone has a transformer architecture, containing a CNN encoder for feature extraction, a transformer encoder, a transformer decoder, and a CNN decoder for mapping the deep features into restored images.
With the self-attention mechanism, the transformer can separate the generalized prior from the corrupted images.
Different from conventional transformer~\cite{dosovitskiy2020image}, the decoder of our transformer takes additional prior queries, which comes from the prior learning module.

\textbf{CNN encoder and CNN decoder.} The CNN encoder consists of two $3 \times 3$ convolutional layers. The RGB image, $I \in \mathbb{R}^{3 \times H \times W}$, is the input of the CNN encoder, which generates a feature map $f_{e} \in \mathbb{R}^{64 \times H \times W}$ with 64 channels and with the same resolution as $I$. The CNN decoder also consists of two $3 \times 3$ convolutional layers. It generates a reconstructed image $O \in \mathbb{R}^{3 \times H \times W}$. 

\textbf{Transformer encoder.} The feature map $f_{e}$ is firstly flattened into small patches $\{ f^{1}_{e}, f^{2}_{e},..., f^{N}_{e} \}$, where $f^{i}_{e} \in \mathbb{R}^{64P^{2}}(i=1,2,...,N)$, $N=\frac{H W}{P^{2}}$ is the total patch number and $P$ is the patch size.
A learnable position encoding $PE_{i}$ with the same size of $f^{i}_{e}$ is added to $f^{i}_{e}$ and the sum (denoted as $x_{i}$) is sent into the transformer encoder. The transformer encoder has $n$ transformer blocks ($n=1$ in this work), each having a multi-head self-attention module and a feed forward network.
The process of the transformer encoder can be formulated as,
\begin{equation}
\begin{array}{l}
x^{\prime}=\operatorname{MSA}\left(\mathrm{LN}\left(x\right), \mathrm{LN}\left(x\right), \mathrm{LN}\left(x\right)\right)+x \\
o_{e}=\operatorname{FFN}\left(\mathrm{LN}\left(x^{\prime}\right)\right)+x^{\prime}, \\
\end{array}
\end{equation}
where MSA, FFN, and LN denote the multi-head self-attention module, feed forward network, and linear layer in the conventional transformer~\cite{vaswani2017attention}, respectively. $x=\left[x_{1}, x_{2}, \ldots, x_{N}\right]$ and $o_{e}=\left[o_{e_{1}}, o_{e_{2}}, \ldots, o_{e_{N}}\right]$ are the input and the output with the same size, respectively.

\textbf{Prior learning module.}
The prior learning module (PLM) aims at providing additional prior queries to the transformer decoder. 
As shown in Fig. \ref{network_ipt}, PLM encodes an image into a feature map, and it can be formulated as $f_{n}=\mathrm{G_{n}}(I_{gt})$, where $f_{n}$ is a $64 \times H \times W$ feature map representing the deep natural image features and $\mathrm{G_{n}}$ is a VGG19 network to extract image features.
Then $f_{n}$ is flattened into a series of patches $\left[f_{n}^{1}, f_{n}^{2}, \ldots, f_{n}^{N}\right]$ and combined with learnable parameters $\left[e_{1}, e_{2}, \ldots, e_{N}\right]$ as follows,

\begin{equation}
    Q = \left[e_{1}+f_{n}^{1}, e_{2}+f_{n}^{2}, \ldots, e_{N}+f_{n}^{N}\right],
\end{equation}
where $Q$ have the same length as $o_{e}$.

\textbf{Transformer decoder.}
The transformer decoder has a similar architecture as the transformer encoder except for an additional input of the prior queries $Q$ (Similar with the object queries in~\cite{Dai2021UPDETRUP}). In this paper, we use one transformer decoder block that consists of two multi-head self-attention (MSA) layers and one feed forward network (FFN).
The transformer decoder is
formulated as,
\begin{equation}
\begin{array}{l}
y=\operatorname{MSA}\left(\mathrm{LN}\left(o_{e}\right)+Q, \mathrm{LN}\left(o_{e}\right)+Q, \mathrm{LN}\left(o_{e}\right)\right)+o_{e} \\
y^{\prime}=\operatorname{MSA}\left(\mathrm{LN}\left(y\right)+Q, \mathrm{LN}\left(o_{e}\right), \mathrm{LN}\left(o_{e}\right)\right)+y \\
o_{d}=\mathrm{FFN}\left(\mathrm{LN}\left(y^{\prime}\right)\right)+y^{\prime}, \\
\end{array}
\end{equation}
where $o_{d}=\left[o_{d_{1}}, o_{d_{2}}, \ldots, o_{d_{N}}\right]$ denotes the outputs of the transformer decoder. And then these patches are reshaped into $f_{d}$ with the size of $64 \times H \times W$. 

\begin{figure*}[t!]
		\centering
		\includegraphics[width=0.90\textwidth]{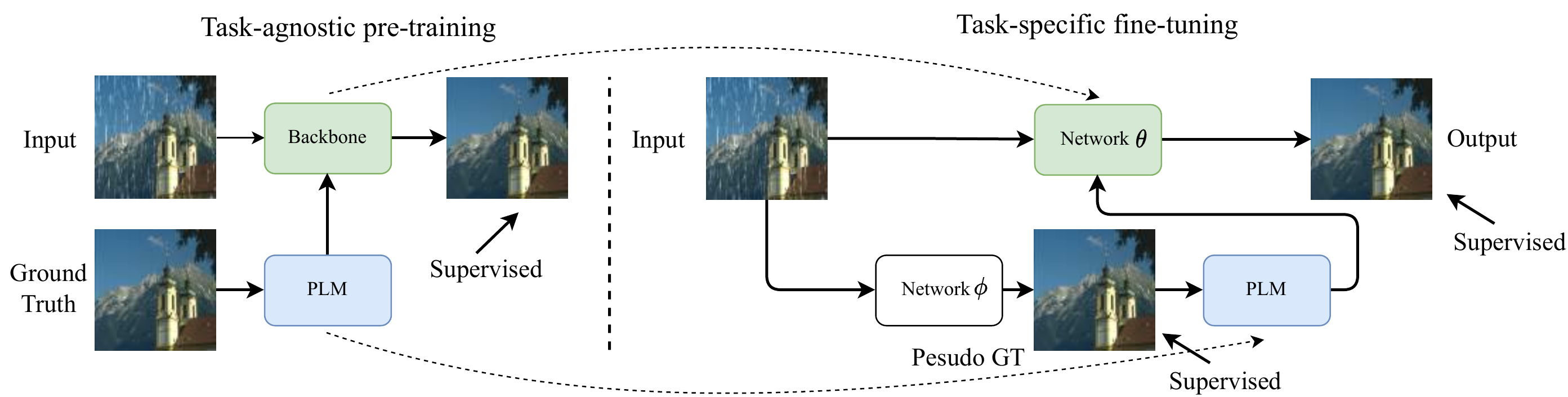}
		\caption{The optimization procedure of TAPE. TAPE contains two stages: task-agnostic pre-training and task-specific fine-tuning. The dotted line means that the network trained in the previous stage is used to initialize the network in the next stage.}
		\label{network_process}
\end{figure*}

\subsection{Optimization}
\label{sec:GDP}

\subsubsection{The Training and Fine-tuning Pipeline}
\ 

From Sec. 3.2, in our network design, PLM should extract statistics from the non-degraded (Ground truth) image and use the statistics to assist the main network for image restoration.
However, ground truth is unavailable during the inference (test) stage. So PLM cannot extract statistics from it. To compensate, we train an auxiliary module (backbone $\phi$ in Fig.~\ref{network_process}) to generate a pseudo GT from the degraded input. The pseudo GT, not being perfect, depicts the property of a non-degraded image to some extent. The pseudo GT is then fed into PLM for extracting general image priors, and the priors assist the main network (backbone $\theta$) to generate the final output.
As shown in Fig. \ref{network_process}, TAPE-Net has two stages: task-agnostic pre-training and task-specific fine-tuning.

In the task-agnostic pre-training, multiple low-level vision tasks are trained together, using corresponding datasets $\{D_{1},...,D_{m}\}$, where $D_{i}, (i=1,2,...,m)$ represents the dataset for task $i$. 
In each iteration, a pair of images (a corrupted image $I_{cor}$ and its ground truth $I_{gt}$) are selected from one dataset $D_{i}$.
The ground truth $I_{gt}$ is sent into PLM to learn the prior queries, which are then sent to the backbone for end-to-end training.
The combination of the $L_1$ loss and the proposed pixel-wise contrastive loss (see section \ref{pcloss}) is used to optimize the network (the weighting parameter is  for the latter).
Due to the task-agnostic pre-training, both the backbone and PLM are well optimized.

As shown in the right part of Fig. \ref{network_process}, in the task-specific fine-tuning, because we cannot take the ground truth as input as discussed above, we use an auxiliary module (network $\phi$ in Fig. 3)  to generate a pseudo GT from the degraded input (Note that in our paper, the network $\phi$ can be any neural network or the backbone borrowed from the pre-training stage. For faster convergence, we use the network pre-trained in the first stage).
And then, the pseudo GT generated by the network $\phi$ is served as the input of pre-trained PLM.
The PLM outputs the prior queries and help the pre-trained network $\theta$ to predict better final result. With the estimated pseudo ground truth, PLM can capture the natural image priors better.
There are two ways to optimize the parameters in the fine-tuning stage, namely: 1) The network $\phi$ is fine-tuned by loss between pseudo GT and GT firstly, and then fixed when fine-tuning other networks; 2) All components are fine-tuned simultaneously. 
In the supplementary material, we will show that these two optimization methods lead to similar performance.

\subsubsection{Pixel-wise Contrastive Loss}
\ 

\label{pcloss}

PLM aims to estimate the distribution of natural patches. However, due to the limited amount of training data, learning from the loss between prediction and GT (unary term) is insufficient for accurate estimation.
Inspired by some self-supervised learning for high-level semantics (\textit{e.g.} MoCo~\cite{KaimingHe2020MomentumCF} and SimCLR~\cite{TingChen2020ASF}) and image to image translation method~\cite{KyungjuneBaek2021RethinkingTT,TaesungPark2020ContrastiveLF}, we propose a pixel-wise contrastive loss to offer another cue (binary terms) of estimation -- the distance between the features of $I_{d}$ and $I_{gt}$ (from the same location) shall be smaller than that between features from different locations.

\begin{figure}[t]
		\centering
		\includegraphics[width=0.88\textwidth]{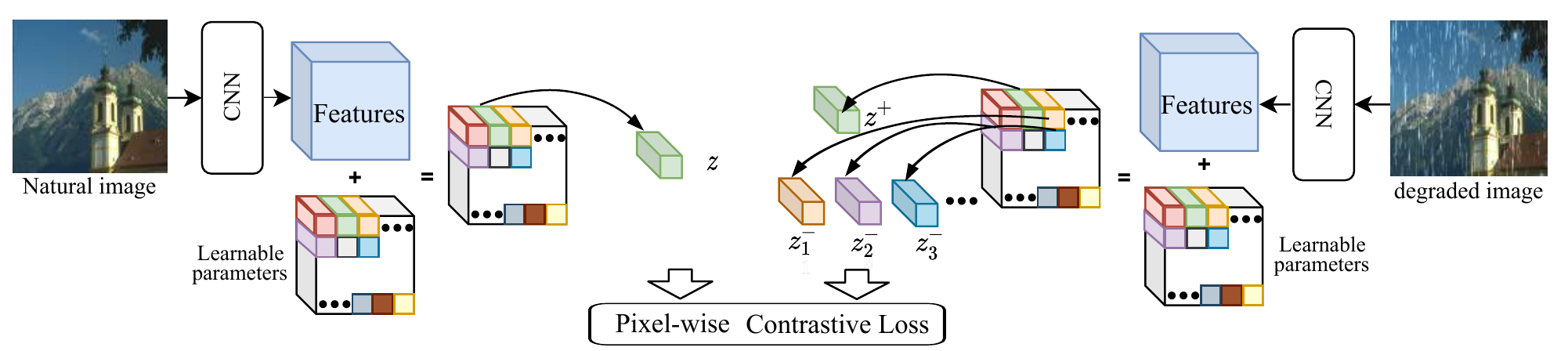}
		\caption{The use of pixel-wise contrastive loss. $z$ is selected from the prior queries of the natural image as `query'. And $z^{+}$ and $z^{-}_{j}$ are selected as the `positive' and `negative' elements in the contrastive loss, respectively.}		\label{fig: closs}
\end{figure}

As shown in Fig. \ref{fig: closs}, in the task-agnostic pre-training stage, the degraded image $I_{d}$ and the natural image $I_{gt}$ are put into PLM, then $Q^{d}$ and $Q^{gt}$ are obtained as described in Sec. 3.2. 
We aim at minimizing the distance between the features of $I_{d}$ and $I_{gt}$ from the same location while maximizing the distance between features from different locations.
For example, in Fig. \ref{fig: closs}, the roof without rain should be more closely associated with the roof contaminated by the rain than the other patches of the rainy input, such as other parts of the house or the blue sky.

Suppose that $q^{d}_{i}$ is selected from $Q^{d}=\{q^{d}_{1},\ q^{d}_{2}, . . . ,\ q^{d}_{N}\}$ as the `query' element in the contrastive loss. $q^{gt}_{i}$ and $q^{gt}_{j_{1}}$, $q^{gt}_{j_{2}}$,..., $q^{gt}_{j_{m}}$ are selected from $Q^{gt}=\{q^{gt}_{1},\ q^{gt}_{2}, . . . ,\ q^{gt}_{N}\}$ as the `positive' and `negative' elements in the contrastive loss, respectively.
Thus, the contrastive loss is formulated as, 
\begin{equation}
\mathcal{L}= \sum_{t=1}^{T} \ell_{t}\left(q^{d}_{i}, q^{gt}_{i}, q^{gt}_{j}\right),
\end{equation}

\begin{equation}
 \hspace{-2mm}
 \begin{array}{c}
\ell\left(q^{d}_{i}, q^{gt}_{i}, q^{gt}_{j}\right)\!=\!  -\log \left[\frac{\exp \!\left(q^{d}_{i} \cdot q^{gt}_{i} / \tau\right)}{\exp \!\left(q^{d}_{i} \cdot q^{gt}_{i} / \tau\right)+\sum_{k=1}^{m} \exp \! \left(q^{d}_{i} \cdot q^{gt}_{j_{k}} / \tau\right)}\right],
\end{array}
\end{equation}
where $T=256$ is the feature number we randomly choose each time and the temperature $\tau$ is set to $0.07$. The negative sample number $m$ is set to 256 in our work.

\subsection{Relationship to Prior Work}
1) Compared with recent multiple degradation prior learning methods (e.g., IPT and EDT), our TAPE formulates generalized image priors, which means that our method can generalize well to the pre-training-unknown tasks.
Recently, AirNet~\cite{AirNet} also has the ability to generalize to unknown tasks, but their learned representation contains the degraded information instead of normal image information through contrastive learning.
2) Different from the methods~\cite{gu2020image,pan2020exploiting,bau2020semantic,Wang_2021_CVPR} where the learned image priors are hidden in the parameters of the generator, the learned prior of our model is explicit.
It makes easy for our method to disentangle the unwanted noise from the image contents in some complex image restoration cases.

\section{Experiments and Analysis}

In this section, we evaluate the performance of TAPE on several low-level vision tasks and conduct an ablation study. 

\subsection{Tasks and Datasets}
For pre-training, we use five datasets, each for one type of degradation. We also test on four more datasets for four tasks that are unknown in the pre-training stage. 
For both training and testing, we resize images into the resolution of $256 \times 256$, and then crop them into $64 \times 64$ patches for balancing the training procedure with different data sizes.
Note that same resizing and cropping operations are also adopted for other models for a fair comparison.
The evaluated tasks include denoising, deraining, deraindrop, demoireing, desnowing, shadow removal, super resolution and deblurring. As shown in Table \ref{tab:expresults1} for details, 
the used datasets are:
SIDD~\cite{abdelhamed2018high} for denoising, Rain200H and Rain200L ~\cite{yang2017RainRemoval} for deraining, Raindrop800~\cite{qian2018attentive} for deraindrop, TIP2018~\cite{sun2018moire} for demoireing\footnote{\footnotesize{We select a subset of TIP2018 and Snow100K with 10000 training image pairs and 200 test pairs; 10000 training image pairs and 500 test pairs, respectively.}}, Snow100K~\cite{liu2018desnownet} for desnowing, ISTD~\cite{wang2018STCGAN} for shadow removal, DIV2K~\cite{Agustsson_2017_CVPR_Workshops} for super resolution and REDS~\cite{Nah_2019_CVPR_Workshops_REDS} for deblurring.

\begin{table*}[t]
  \begin{center}
  \caption{Datasets' statistics (number of training and testing images) and quantitative comparison for two models (in terms of PSNR (dB)).}
  \label{tab:expresults1}
  \resizebox{12.4cm}{!}{
  \begin{tabular}{l|ccccc|cccc}
    \toprule
      Dataset & Rain200L & Rain200H &Raindrop800&SIDD& TIP2018  &Snow100K  &ISTD & DIV2K & REDS \\
    \midrule 
     \#Train/test images & 1800/200 & 1800/200 & 800/60 & 96000/1280& 10000/200 & 10000/500&1330/540&800/100 & 24000/3000\\

    \midrule 

    Baseline &31.72 &22.81&26.85&37.41&26.77&25.42 & 26.28 &31.25 & 32.46\\
    
    TAPE-Net (Ours) &
     33.17 &
     23.84 &
     27.69 &
     37.90 &
     27.52 &
     26.33 &
     26.57 &
     31.66 &
     32.94 \\
     PSNR Gain & +1.45 & +1.03& +0.84& +0.49& +0.75& +0.91& +0.29& +0.41& +0.48\\

    \bottomrule
  \end{tabular}}
  \end{center}
\end{table*}

\begin{figure*}[t]
		\centering
		\includegraphics[width=0.99\textwidth]{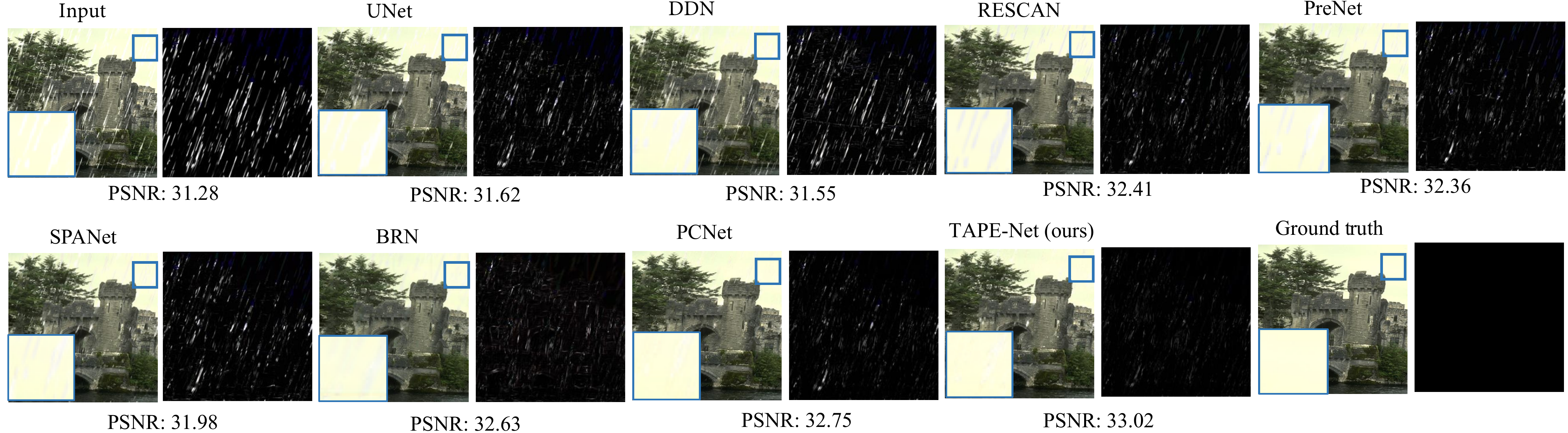}
		\caption{Visual deraining comparison among our methods and SOTA deraining methods on Rain200L. The differences between the output and the ground truth are shown followed the predicted results.}
		\label{fig:rain100l}
\end{figure*}

\subsection{Implementation Details}
\textbf{Pre-training.}
We use one Nvidia Tesla V100 GPU to train our model using the Adam optimizer for $60 \times 24000$ iterations on the mixture of the five dataset (SIDD, Rain200L, Rain200H, Raindrop800, and TIP2018). 
The initial learning rate is set as 0.0002 and decayed to 0.0001 in the $20 \times 24000$th iteration with batch size 4.
In each iteration, we first randomly choose a dataset, from which one clean-corrupted image pair is randomly selected. 
\textbf{Fine-tuning.}
After pre-training on all the datasets, we fine-tune the model on each desired task (\textit{e.g.}, denoising). %
TAPE-Net is fine-tuned with 200 epochs and a learning rate of 2e-4 for task-specific fine-tuning.

\subsection{Pre-training \& Generalization Ability}
In this subsection, we illustrate that our method has good generalization performance on both pre-training-known tasks and pre-training-unknown tasks. 

\textbf{Pre-training-known tasks, corresponding data.} 
To illustrate the effectiveness of our task-agnostic pre-training, we compare our pre-trained model with the model without pre-training (denoted as `Baseline' in Table \ref{tab:expresults1}). 
The TAPE-Net improve the PSNR by 1.45dB, 1.03dB, 0.84dB, 0.49dB and 0.75dB on the Rain200L, Rain200H, Raindrop800, SIDD, and TIP2018 dataset, respectively. Please note that in image restoration, an 0.5dB improvement is usually considered significant.
It demonstrates the effectiveness of the pre-training and the superiority of our model.

\textbf{Pre-training-known tasks, different data.} And we also do experiments to explore the good generalization performance of our pre-trained model when transferred to different distributions of data in the pre-training-known task.
Our experiments show that pre-training with synthetic Gaussian noises helps to restore the images corrupted by real noises (see Table \ref{tab:noise_abstudy}).
In practice, in the task-agnostic pre-training, we use ground truth in SIDD with added synthetic Gaussian noises as input, without touching the real noise images on the SIDD dataset.
Then the pre-trained model is fine-tuned on SIDD (using real-noise/non-noise image pairs).
The PSNRs increase by 0.19dB, and 0.31dB respectively when the $\sigma$ of the added Gaussian noise (Sampled from $\mathcal{N}(0,\sigma)$) are in [5,20], and [1,50], respectively. 
When the range of the added Gaussian noise is larger, the generalization ability of the model in the pre-training stage is stronger, and the performance in the fine-tuning stage is better.

\textbf{Pre-training-unknown tasks.} To illustrate the generalization ability of our model, we conduct several experiments on the tasks that are unknown to the pre-training stage.
In practice, we fine-tune the pre-trained model on four new low-level vision tasks: desnowing, shadow removal, super resolution and deblurring. %
As shown in Table~\ref{tab:expresults1}, compared with the none-pre-trained model, TAPE-Net improves the PSNR by 0.91dB, 0.29dB, 0.41dB and 0.48dB, which demonstrates that the pre-trained model can capture more useful information and features from natural images.
Learning task-agnostic priors and pixel-wise contrastive loss on pre-training stage can help the performance of fine-tuning on the unknown tasks.

\begin{figure*}[t]
		\centering
		\includegraphics[width=0.99\textwidth]{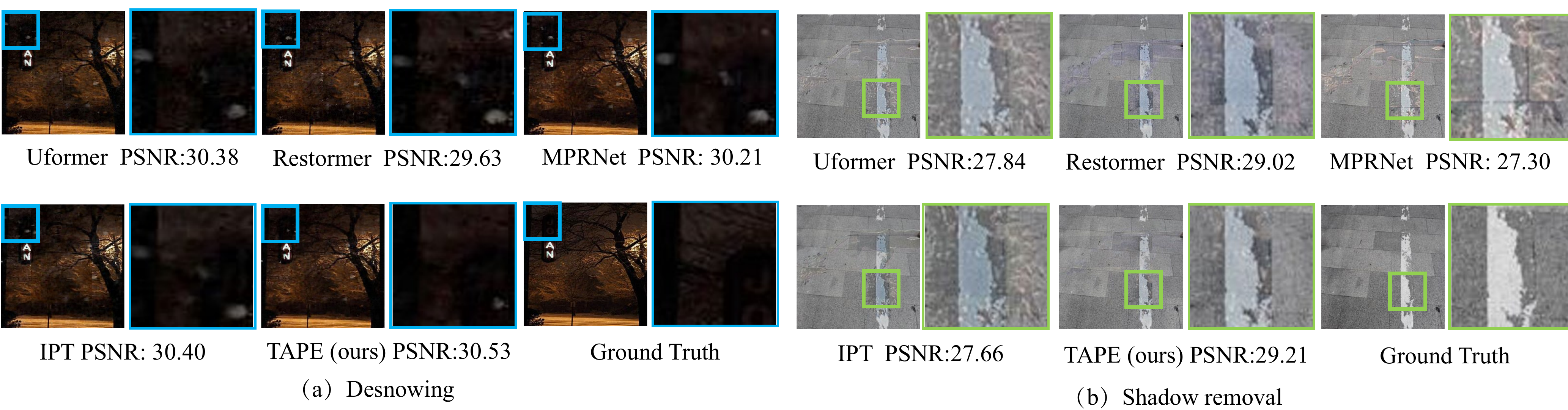}
		\caption{Visual desnowing and shadow removal comparison among our methods and SOTA general image restoration methods on Snow100K and ISTD.}
		\label{fig:sotacompare}
\end{figure*}

\begin{table}[t]
  \begin{center}
  \caption{Quantitative comparison with 4 SOTA general image restoration methods with similar model sizes (in terms of PSNR (dB)).}
  \label{tab:sotatrans}
  \resizebox{8.9cm}{!}{
  \begin{tabular}{lcccccc}
    \toprule
         & Model size & Snow100K & ISTD  & Rain200L  &SIDD & Raindrop800 \\
    \midrule
    IPT & 2.51M & 26.26 &26.34&32.67 & 38.80 & 27.86\\
    Restormer & 0.93M &26.80  & 26.42&33.61  & 38.91 & 27.98\\
    UFormer & 0.97M &26.50 & 26.27& 32.66 &38.84  &27.70\\
    MPRNet & 1.11M & 26.30 & 26.23 & 33.30 & 38.89 & 28.13 \\
    TAPE-Swin (Ours)  &0.97M & \textbf{26.93} &26.61 & \textbf{34.46} & 38.98& 29.15    \\
   TAPE-Restormer (Ours) & 1.07M & 26.91 & \textbf{26.65} & 34.28 & \textbf{39.01} & \textbf{29.18} \\

    \bottomrule
  \end{tabular}}
  \end{center}
\end{table}

\subsection{Comparisons with State-of-the-Arts}

\begin{figure*}[t]
		\centering
		\includegraphics[width=0.98\textwidth]{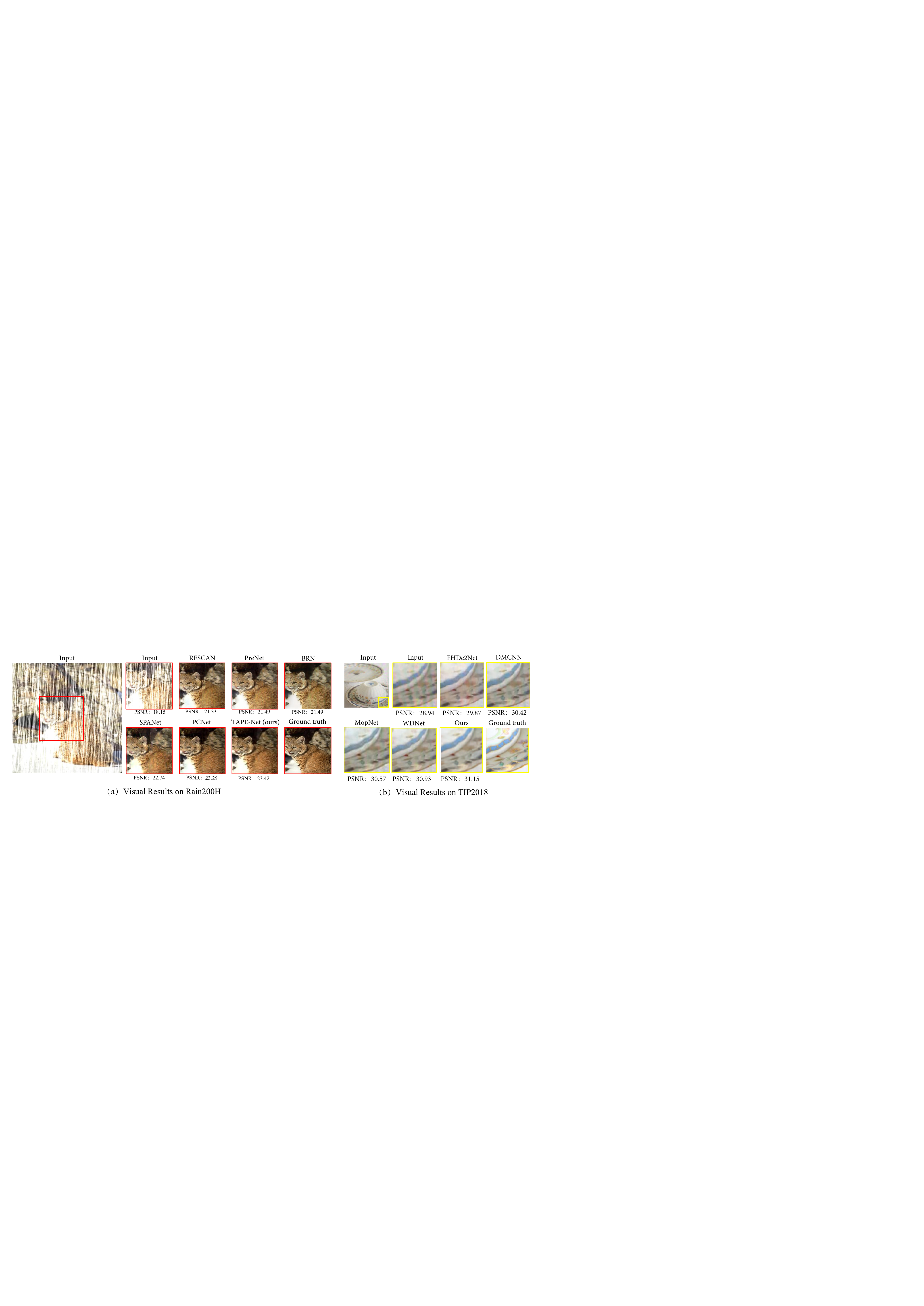}
		\caption{Visual deraining comparison (a) and  visual demoireing comparison (b) among our methods and task-specific methods.}
		\label{fig:rain100h}
\end{figure*}

In this subsection, the comparison between our methods and the very recent SOTA image restoration methods are shown in Table \ref{tab:sotatrans}.  we compare our methods (TAPE-swin and TAPE-restormer) with the 4 SOTA methods (IPT~\cite{chen2020pre}, Restormer~\cite{Zamir2021Restormer},
UFormer~\cite{wang2021uformer}, MPRNet~\cite{Zamir2021MPRNet}) with the similar model parameters in 3 pre-training-known tasks and 2 pre-training-unknown tasks. The visual results on two pre-training-unknown tasks (desnowing and shadow removal) are shown in Fig.~\ref{fig:sotacompare}. Our method removes the artifacts more thoroughly and retains more details.
In the supplementary material, we compare our TAPE-Net with 12 state-of-the-art task-specific methods, including deraining methods (DDN~\cite{fu2017removing}, SPANet~\cite{wang2019spatial}, RESCAN~\cite{li2018recurrent}, PreNet~\cite{ren2019progressive}, BRN~\cite{ren2020single},SPDNet~\cite{QiaosiYi2021StructurePreservingDW} and PCNet~\cite{2021pcnet}); demoireing methods ( DMCNN~\cite{sun2018moire}, MopNet~\cite{he2019mop}, FHDe2Net~\cite{he2019mop}, HRDN~\cite{yang2020high}, WDNet~\cite{liu2020wavelet} and MBCNN \cite{zheng2021learning}); and denoising methods (DnCNN~\cite{zhang2017beyond}, FFDNet~\cite{zhang2018ffdnet}, RDN~\cite{zhangrdn2020}, and SADNet~\cite{chang2020spatial}) on PSNR.
Qualitative results on deraining are shown in Figs. \ref{fig:rain100l} and \ref{fig:rain100h}(a), showing that our methods get cleaner images and recover more details.
The visual demoireing results are shown in Fig. \ref{fig:rain100h}(b). Ours can remove moire patterns successfully and restore the underlying clean image.

In order to show the performance of our method on pre-training-unknown tasks, we compare ours with the existing SOTA multi-task pre-training method, IPT~\cite{chen2020pre}.
We fine-tuned and tested the official pre-trained model of IPT on three unseen tasks, namely, desnowing, shadow removal and deblurring.
We compared the gain of PSNR with and without pre-training, because the training patches and model size of IPT and ours are different.
The pre-training stage of IPT boosts PSNR by 0.07dB, 0.19dB and 0.08dB, respectively, while the improvements of TAPE (our method) are 0.91dB,  0.29dB and 0.48dB, larger than that of IPT.

\begin{table}[t]
  
\begin{center}
  \caption{Quantitative comparison on SIDD. It shows the good generalization performance of TAPE-Net when transferred to different distributions of data in the pre-training-known task.}
  \label{tab:noise_abstudy}
  \resizebox{8.8cm}{!}{
  \begin{tabular}{cccc}
    \toprule
      TAPE-Net & W/o pre-train & Pre-train ($\sigma$ $\in$ [5,20]) &  Pre-train ($\sigma$ $\in$ [1,50])\\
    \midrule 
     PSNR& 37.41 & 37.60 & 37.72 \\
    \bottomrule
  \end{tabular}}
  \end{center}
\end{table}

\begin{table}[t!]
  \begin{center}
  \caption{Ablation study of the impact of the multi-task pre-training, which illustrates that adding more tasks will not harm the performance. }
  \label{tab:abstudy_num}
  \resizebox{7.8cm}{!}{
  \begin{threeparttable}
  \begin{tabular}{cccc|c }
    \toprule
      & Raindrop800 & Rain200L & TIP2018&Snow100K\\
    \midrule 
   RD+RL+RH &27.71 & 33.20 &27.56 & 26.17\\
        RD+TIP+RL+RH & 27.70& 33.18 &27.54 &26.26 \\

    RD+TIP+RL+RH+S  &27.69 & 33.17 &27.52 &26.33 \\
    \bottomrule
    
  \end{tabular}

  \end{threeparttable}}
   
  \end{center}
\end{table}

\subsection{Ablation Study}

\begin{table}[t]
  \begin{center}
  \caption{The ablation study of the importance of each pre-trained part on the Rain200H and the Raindrop-TestB dataset. `\textbf{-}' and `$\checkmark$' in the first three columns mean that the model parameters are randomized and pre-trained respectively before the finetuning. `\ding{55}' means the corresponding part does not exist.}
  \resizebox{12.3cm}{!}{
  \setlength{\tabcolsep}{0.12cm}
  \begin{tabular}{c|ccccccc}
    \toprule
    Name & Depth &network $\phi$ & network $\theta$ & PLM  & Rain200H & Raindrop-TestB & Model size (M)\\
    \midrule 
    Baseline & 3 &  \ding{55} & \textbf{-} & \ding{55} & 25.57 & 26.13 & 0.76\\
    
    With no-pre-trained all parts&2 & \textbf{-} &\textbf{-} &\textbf{-}  & 25.46& 25.96 & 1.12\\

    \midrule
    With no-pre-trained $\phi$& 2 & \textbf{-} &  \checkmark & \checkmark & 26.15 &26.39 &1.12\\
    
    With no-pre-trained $\theta$ & 2 &  \checkmark&  \textbf{-} &  \checkmark& 25.96 & 26.35& 1.12\\
    
     With no-pre-trained $\theta$ and PLM  & 2 & \checkmark & \textbf{-}& \textbf{-} &  25.73 & 26.17 & 1.12\\
     
     \midrule
     Full model & 2 &\checkmark  &\checkmark  &\checkmark  & \bf{26.18} & \bf{26.41} & 1.12\\
     
    \bottomrule
  \end{tabular}}
  
  \label{tab:abstd}
  \end{center}
\end{table}

\textbf{Impact of the multi-task pre-training.}
We do ablation study to analyze the effect of the number of datasets in the pre-training.
We pre-train our models on fewer datasets and compare with our original models.
As shown in Table \ref{tab:abstudy_num}, RD, TIP, RL, RH, and S mean Raindrop800, TIP2018, Rain200L, Rain200H, and SIDD datasets respectively. `+' means we use these datasets in the pretraining stage.
We do the experiments on three pre-trained-known datasets and the maximum PSNR difference is 0.04 dB. 
This PSNR difference is within a controllable error range.
Increasing the dataset in the pre-training is meaningful. Compared with pre-training with 3 datasets, pre-training with 5 datasets increases the PSNR by 0.16dB on Snow100K.

\textbf{Importance of pre-training of each component.}
In order to verify which component is more important with pre-training (network $\phi$, network $\theta$ or PLM), we randomize the weights of corresponding parts of TAPE-Net before the fine-tuning stage.
As shown in Table \ref{tab:abstd}, we do ablation study on Rain200H and Raindrop-TestB.
All the modified models in this table have the same backbone, TAPE-Swin.
Compared `With no-pre-trained $\phi$' and the `Full model', we can see that the PSNRs drop slightly (0.03dB and 0.02dB).
With no-pre-trained $\theta$ effects more than the network $\phi$ (the PSNRs drop by 0.22dB and 0.06dB).
`Without pre-training the network $\theta$ and PLM' also makes the PSNRs drop a lot.
Thus, the last four lines in Table~\ref{tab:abstd} demonstrate that the effectiveness of pre-training of $\theta$ and PLM.
The first two lines in Table~\ref{tab:abstd} show that even if the model size is larger than Baseline, the performance is not good without pre-training.

The ablation studies about the pixel-wise contrastive loss, optimization methods in task-specific fine-tuning can be see in the supplementary material.

\subsection{Visualization Results}

\begin{figure*}[t!]
		\centering
		\includegraphics[width=0.92\textwidth]{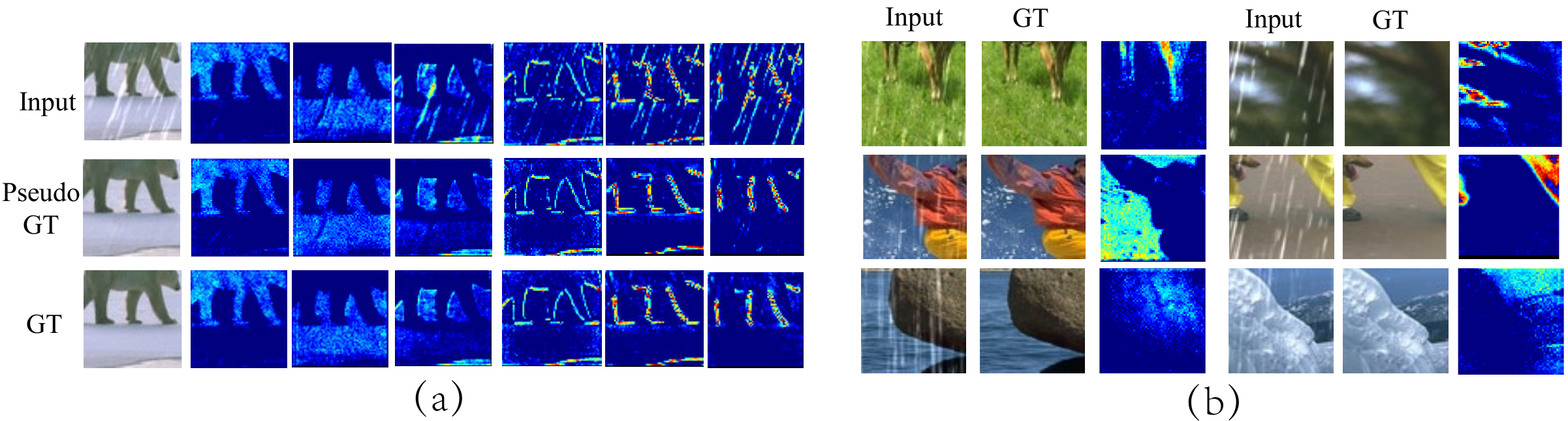}
		\caption{The visualization of the PLM's output, $Q$. (a) The features of the pseudo GT and GT are very similar, but most of the features of input have the features of degraded objects. It means that the features of the pseudo GT is useful and can help the image restoration of the backbone. (b) The last lines are one of the predicted results of PLM.}
		\label{visualq}
\end{figure*}

To validate that PLM learns useful and meaningful features with our proposed pipeline, we visualize the features learned by PLM on the deraining task.
We put the input, pseudo GT and GT into PLM to get their respective output $Q$.
As shown in Fig.~\ref{visualq}(a), in the output features, some channels tend to preserve the information of shapes (the first three feature maps) and edges (the last three feature maps).
The features of the pseudo GT and GT are very similar, but most of the features of input have the features of degraded objects.
It means that the features of the pseudo GT is useful and can help the image restoration of the backbone. 
From Fig.~\ref{visualq}(b), we can see that with the help of pre-training,
the PLM module can correlate the information of similar textures or patches
from a long distance. Thus, the transformer decoder of the backbone can utilize
these long-distance similar areas/patches to restore the image.

\section{Conclusions and Limitations}

\textbf{Conclusions.}
In this paper, we address the possibility and importance of learning task-agnostic and generalized image prior. We propose a pipeline named TAPE to learn task-agnostic prior embedding for image restoration. 
TAPE has two stages: task-agnostic pre-training and task-specific fine-tuning.
Our task-agnostic training strategy is able to learn generalized natural image prior.
It has good generalization performance when faced with pre-training-unknown tasks.

\textbf{Limitations.} Although our method shows generalized ability on a few image restoration tasks, the learned statistics in PLM are still difficult to explain, either in theory or by visualization results. Without these proofs, the PSNR and SSIM numbers are only side evidences of the effectiveness of the task-agnostic priors. In the future, we will continue exploring the possibility of disentangling the task-agnostic priors as well as finding more essential ways to evaluate cross-task image restoration. The pipeline may make the training time longer than the baseline. And the performance of TAPE on mixed degradation tasks needs to be explored.

\textbf{Acknowledgements.} This work was supported by the National Natural Science Foundation of China under Contract 61836011 and 62021001. It was also supported by the GPU cluster built by MCC Lab of Information Science and Technology Institution, USTC.

\bibliographystyle{splncs04}
\bibliography{egbib}

\begin{thebibliography}{10}
\providecommand{\url}[1]{\texttt{#1}}
\providecommand{\urlprefix}{URL }
\providecommand{\doi}[1]{https://doi.org/#1}

\bibitem{abdelhamed2018high}
Abdelhamed, A., Lin, S., Brown, M.S.: A high-quality denoising dataset for
  smartphone cameras. In: CVPR, 2018

\bibitem{Agustsson_2017_CVPR_Workshops}
Agustsson, E., Timofte, R.: Ntire 2017 challenge on single image
  super-resolution: Dataset and study. In: CVPRW, 2017

\bibitem{babacan2008variational}
Babacan, S.D., Molina, R., Katsaggelos, A.K.: Variational bayesian blind
  deconvolution using a total variation prior. TIP, 2008

\bibitem{KyungjuneBaek2021RethinkingTT}
Baek, K., Choi, Y., Uh, Y., Yoo, J., Shim, H.: Rethinking the truly
  unsupervised image-to-image translation. In: International Conference on
  Computer Vision (ICCV, 2021)

\bibitem{bau2020semantic}
Bau, D., Strobelt, H., Peebles, W., Zhou, B., Zhu, J.Y., Torralba, A., et~al.:
  Semantic photo manipulation with a generative image prior. arXiv preprint
  arXiv:2005.07727  (2020)

\bibitem{AndrewBrock2018LargeSG}
Brock, A., Donahue, J., Simonyan, K.: Large scale gan training for high
  fidelity natural image synthesis. In: ICLR, 2018

\bibitem{NicolasCarion2020EndtoEndOD}
Carion, N., Massa, F., Synnaeve, G., Usunier, N., Kirillov, A., Zagoruyko, S.:
  End-to-end object detection with transformers. In: ECCV,2020

\bibitem{chan2020glean}
Chan, K.C., Wang, X., Xu, X., Gu, J., Loy, C.C.: Glean: Generative latent bank
  for large-factor image super-resolution. arXiv preprint arXiv:2012.00739
  (2020)

\bibitem{chang2020spatial}
Chang, M., Li, Q., Feng, H., Xu, Z.: Spatial-adaptive network for single image
  denoising. ECCV, 2020

\bibitem{chen2020pre}
Chen, H., Wang, Y., Guo, T., Xu, C., Deng, Y., Liu, Z., Ma, S., Xu, C., Xu, C.,
  Gao, W.: Pre-trained image processing transformer. CVPR, 2021

\bibitem{Chen_2019_CVPR}
Chen, L., Fang, F., Wang, T., Zhang, G.: Blind image deblurring with local
  maximum gradient prior. In: CVPR, 2019

\bibitem{TingChen2020ASF}
Chen, T., Kornblith, S., Norouzi, M., Hinton, G.E.: A simple framework for
  contrastive learning of visual representations. In: ICML, 2020

\bibitem{chen2020jstasr}
Chen, W.T., Fang, H.Y., Ding, J.J., Tsai, C.C., Kuo, S.Y.: Jstasr: Joint size
  and transparency-aware snow removal algorithm based on modified partial
  convolution and veiling effect removal. In: ECCV, 2020

\bibitem{chen2021all}
Chen, W.T., Fang, H.Y., Hsieh, C.L., Tsai, C.C., Chen, I., Ding, J.J., Kuo,
  S.Y., et~al.: All snow removed: Single image desnowing algorithm using
  hierarchical dual-tree complex wavelet representation and contradict channel
  loss. In: CVPR, 2021

\bibitem{XiangyuChen2022ActivatingMP}
Chen, X., Wang, X., Zhou, J., Dong, C.: Activating more pixels in image
  super-resolution transformer. arXiv preprint arXiv:2205.04437  (2022)

\bibitem{chen2013generalized}
Chen, Y.L., Hsu, C.T.: A generalized low-rank appearance model for
  spatio-temporally correlated rain streaks. In: ICCV, 2013

\bibitem{cun2019ghostfree}
Cun, X., Pun, C.M., Shi, C.: Towards ghost-free shadow removal via dual
  hierarchical aggregation network and shadow matting gan. In: AAAI, 2020

\bibitem{dai2021wavelet}
Dai, T., Li, W., Cao, X., Liu, J., Jia, X., Leonardis, A., Yan, Y., Yuan, S.:
  Wavelet-based network for high dynamic range imaging. arXiv preprint
  2108.01434  (2021)

\bibitem{Dai2021UPDETRUP}
Dai, Z., Cai, B., Lin, Y., Chen, J.: Up-detr: Unsupervised pre-training for
  object detection with transformers. In: CVPR, 2021

\bibitem{dong2015image}
Dong, C., Loy, C.C., He, K., Tang, X.: Image super-resolution using deep
  convolutional networks. TPAMI, 2015

\bibitem{dosovitskiy2020image}
Dosovitskiy, A., Beyer, L., Kolesnikov, A., Weissenborn, D., Zhai, X.,
  Unterthiner, T., Dehghani, M., Minderer, M., Heigold, G., Gelly, S., et~al.:
  An image is worth 16x16 words: Transformers for image recognition at scale.
  arXiv preprint arXiv:2010.11929, 2020

\bibitem{elhelou2020bigprior}
El~Helou, M., S\"usstrunk, S.: {BIGPrior}: Towards decoupling learned prior
  hallucination and data fidelity in image restoration. arXiv preprint
  arXiv:2011.01406  (2020)

\bibitem{fu2017removing}
Fu, X., Huang, J., Zeng, D., Huang, Y., Ding, X., Paisley, J.: Removing rain
  from single images via a deep detail network. In: CVPR, 2017

\bibitem{AlonaGolts2020UnsupervisedSI}
Golts, A., Freedman, D., Elad, M.: Unsupervised single image dehazing using
  dark channel prior loss. TIP, 2020

\bibitem{gu2020image}
Gu, J., Shen, Y., Zhou, B.: Image processing using multi-code gan prior. In:
  CVPR, 2020

\bibitem{guo2021joint}
Guo, S., Liang, Z., Zhang, L.: Joint denoising and demosaicking with green
  channel prior for real-world burst images. arXiv preprint arXiv:2101.09870
  (2021)

\bibitem{guo2019toward}
Guo, S., Yan, Z., Zhang, K., Zuo, W., Zhang, L.: Toward convolutional blind
  denoising of real photographs. In: CVPR, 2019

\bibitem{hefhde2net}
He, B., Wang, C., Shi, B., Duan, L.Y.: Fhde2net: Full high definition
  demoireing network. ECCV, 2020

\bibitem{he2019mop}
He, B., Wang, C., Shi, B., Duan, L.Y.: Mop moire patterns using mopnet. In:
  ICCV, 2019

\bibitem{KaimingHe2020MomentumCF}
He, K., Fan, H., Wu, Y., Xie, S., Girshick, R.: Momentum contrast for
  unsupervised visual representation learning. In: CVPR, 2020

\bibitem{he2010single}
He, K., Sun, J., Tang, X.: Single image haze removal using dark channel prior.
  TPAMI, 2010

\bibitem{isobe2020video}
Isobe, T., Li, S., Jia, X., Yuan, S., Slabaugh, G., Xu, C., Li, Y.L., Wang, S.,
  Tian, Q.: Video super-resolution with temporal group attention. In: CVPR,
  2020

\bibitem{resunet}
Jha, D., Smedsrud, P.H., Riegler, M.A., Johansen, D., De~Lange, T., Halvorsen,
  P., Johansen, H.D.: Resunet++: An advanced architecture for medical image
  segmentation. In: ISM, 2019

\bibitem{2021pcnet}
Jiang, K., Wang, Z., Yi, P., Chen, C., Lin, C.W.: Pcnet: Progressive coupled
  network for real-time image deraining. In: TIP, 2021

\bibitem{jin2021dc}
Jin, Y., Sharma, A., Tan, R.T.: Dc-shadownet: Single-image hard and soft shadow
  removal using unsupervised domain-classifier guided network. In: ICCV, 2021

\bibitem{kupyn2019deblurgan}
Kupyn, O., Martyniuk, T., Wu, J., Wang, Z.: Deblurgan-v2: Deblurring
  (orders-of-magnitude) faster and better. In: ICCV, 2019

\bibitem{HunsangLee2020UnsupervisedLI}
Lee, H., Sohn, K., Min, D.: Unsupervised low-light image enhancement using
  bright channel prior. IEEE Signal Processing Letters, 2020

\bibitem{levin2009understanding}
Levin, A., Weiss, Y., Durand, F., Freeman, W.T.: Understanding and evaluating
  blind deconvolution algorithms. In: CVPR, 2009

\bibitem{AirNet}
Li, B., Liu, X., Hu, P., Wu, Z., Lv, J., Peng, X.: All-in-one image restoration
  for unknown corruption. In: CVPR (2022)

\bibitem{li2019blind}
Li, L., Pan, J., Lai, W.S., Gao, C., Sang, N., Yang, M.H.: Blind image
  deblurring via deep discriminative priors. IJCV, 2019

\bibitem{li2022sj}
Li, W., Xiao, S., Dai, T., Yuan, S., Wang, T., Li, C., Song, F.: Sj-hd\^{} 2r:
  Selective joint high dynamic range and denoising imaging for dynamic scenes.
  arXiv preprint 2206.09611  (2022)

\bibitem{edtli}
Li, W., Lu, X., Lu, J., Zhang, X., Jia, J.: On efficient transformer and image
  pre-training for low-level vision. In: arXiv preprint 2112.10175

\bibitem{li2018recurrent}
Li, X., Wu, J., Lin, Z., Liu, H., Zha, H.: Recurrent squeeze-and-excitation
  context aggregation net for single image deraining. In: ECCV, 2018

\bibitem{li2020learning}
Li, X., Jin, X., Lin, J., Liu, S., Wu, Y., Yu, T., Zhou, W., Chen, Z.: Learning
  disentangled feature representation for hybrid-distorted image restoration.
  In: ECCV, 2020

\bibitem{li2016rain}
Li, Y., Tan, R.T., Guo, X., Lu, J., Brown, M.S.: Rain streak removal using
  layer priors. In: CVPR, 2016

\bibitem{liang2021swinir}
Liang, J., Cao, J., Sun, G., Zhang, K., Van~Gool, L., Timofte, R.: Swinir:
  Image restoration using swin transformer. In: ICCVW, 2021

\bibitem{liu2020wavelet}
Liu, L., Liu, J., Yuan, S., Slabaugh, G., Leonardis, A., Zhou, W., Tian, Q.:
  Wavelet-based dual-branch network for image demoir{\'e}ing. ECCV, 2020

\bibitem{Liu2022SiamTransZM}
Liu, L., Yuan, S., Liu, J., Guo, X., Yan, Y., Tian, Q.: Siamtrans: Zero-shot
  multi-frame image restoration with pre-trained siamese transformers. In:
  AAAI, 2022

\bibitem{liu2018desnownet}
Liu, Y.F., Jaw, D.W., Huang, S.C., Hwang, J.N.: Desnownet: Context-aware deep
  network for snow removal. TIP, 2018

\bibitem{ZeLiu2021SwinTH}
Liu, Z., Lin, Y., Cao, Y., Hu, H., Wei, Y., Zhang, Z., Lin, S., Guo, B.: Swin
  transformer: Hierarchical vision transformer using shifted windows. CVPR,
  2021

\bibitem{Nah_2019_CVPR_Workshops_REDS}
Nah, S., Baik, S., Hong, S., Moon, G., Son, S., Timofte, R., Lee, K.M.: Ntire
  2019 challenge on video deblurring and super-resolution: Dataset and study.
  In: CVPRW, 2019

\bibitem{Pan_2020_CVPR}
Pan, J., Bai, H., Tang, J.: Cascaded deep video deblurring using temporal
  sharpness prior. In: CVPR, 2020

\bibitem{pan2016blind}
Pan, J., Sun, D., Pfister, H., Yang, M.H.: Blind image deblurring using dark
  channel prior. In: CVPR, 2016

\bibitem{pan2020exploiting}
Pan, X., Zhan, X., Dai, B., Lin, D., Loy, C.C., Luo, P.: Exploiting deep
  generative prior for versatile image restoration and manipulation. In: ECCV,
  2020

\bibitem{TaesungPark2020ContrastiveLF}
Park, T., Efros, A.A., Zhang, R., Zhu, J.Y.: Contrastive learning for unpaired
  image-to-image translation. In: ECCV, 2020

\bibitem{qian2018attentive}
Qian, R., Tan, R.T., Yang, W., Su, J., Liu, J.: Attentive generative
  adversarial network for raindrop removal from a single image. In: CVPR, 2018

\bibitem{ren2020single}
Ren, D., Shang, W., Zhu, P., Hu, Q., Meng, D., Zuo, W.: Single image deraining
  using bilateral recurrent network. TIP, 2020

\bibitem{ren2019progressive}
Ren, D., Zuo, W., Hu, Q., Zhu, P., Meng, D.: Progressive image deraining
  networks: A better and simpler baseline. In: CVPR, 2019

\bibitem{richardson2020encoding}
Richardson, E., Alaluf, Y., Patashnik, O., Nitzan, Y., Azar, Y., Shapiro, S.,
  Cohen-Or, D.: Encoding in style: a stylegan encoder for image-to-image
  translation. arXiv preprint arXiv:2008.00951  (2020)

\bibitem{ronneberger2015u}
Ronneberger, O., Fischer, P., Brox, T.: U-net: Convolutional networks for
  biomedical image segmentation. In: MICCAI, 2015

\bibitem{roth2005fields}
Roth, S., Black, M.J.: Fields of experts: A framework for learning image
  priors. In: CVPR, 2005

\bibitem{rudin1992nonlinear}
Rudin, L.I., Osher, S., Fatemi, E.: Nonlinear total variation based noise
  removal algorithms. Physica D: nonlinear phenomena, 1992

\bibitem{sun2018moire}
Sun, Y., Yu, Y., Wang, W.: Moir{\'e} photo restoration using multiresolution
  convolutional neural networks. TIP, 2018

\bibitem{ulyanov2018deep}
Ulyanov, D., Vedaldi, A., Lempitsky, V.: Deep image prior. In: CVPR, 2018

\bibitem{vaswani2017attention}
Vaswani, A., Shazeer, N., Parmar, N., Uszkoreit, J., Jones, L., Gomez, A.N.,
  Kaiser, L., Polosukhin, I.: Attention is all you need. arXiv preprint
  arXiv:1706.03762, 2017

\bibitem{wang2018STCGAN}
Wang, J., Li, X., Yang, J.: Stacked conditional generative adversarial networks
  for jointly learning shadow detection and shadow removal. CVPR, 2018

\bibitem{wang2019spatial}
Wang, T., Yang, X., Xu, K., Chen, S., Zhang, Q., Lau, R.W.: Spatial attentive
  single-image deraining with a high quality real rain dataset. In: CVPR, 2019

\bibitem{Wang_2021_CVPR}
Wang, X., Li, Y., Zhang, H., Shan, Y.: Towards real-world blind face
  restoration with generative facial prior. In: CVPR, 2021

\bibitem{wang2021uformer}
Wang, Z., Cun, X., Bao, J., Liu, J.: Uformer: A general u-shaped transformer
  for image restoration. arXiv preprint 2106.03106

\bibitem{yang2020learning}
Yang, F., Yang, H., Fu, J., Lu, H., Guo, B.: Learning texture transformer
  network for image super-resolution. In: CVPR, 2020

\bibitem{yang2020high}
Yang, S., Lei, Y., Xiong, S., Wang, W.: High resolution demoire network. In:
  ICIP, 2020

\bibitem{yang2017RainRemoval}
Yang, W., Tan, R.T., Feng, J., Liu, J., Guo, Z., Yan, S.: Deep joint rain
  detection and removal from a single image. In: CVPR, 2017

\bibitem{QiaosiYi2021StructurePreservingDW}
Yi, Q., Li, J., Dai, Q., Fang, F., Zhang, G., Zeng, T.: Structure-preserving
  deraining with residue channel prior guidance. ICCV  (2021)

\bibitem{yu2018crafting}
Yu, K., Dong, C., Lin, L., Loy, C.C.: Crafting a toolchain for image
  restoration by deep reinforcement learning. In: CVPR, 2018

\bibitem{Zamir2021Restormer}
Zamir, S.W., Arora, A., Khan, S., Hayat, M., Khan, F.S., Yang, M.H.: Restormer:
  Efficient transformer for high-resolution image restoration. In: CVPR, 2022

\bibitem{Zamir2021MPRNet}
Zamir, S.W., Arora, A., Khan, S., Hayat, M., Khan, F.S., Yang, M.H., Shao, L.:
  Multi-stage progressive image restoration. In: CVPR (2021)

\bibitem{zeng2020learning}
Zeng, Y., Fu, J., Chao, H.: Learning joint spatial-temporal transformations for
  video inpainting. In: ECCV, 2020

\bibitem{zhang2017beyond}
Zhang, K., Zuo, W., Chen, Y., Meng, D., Zhang, L.: Beyond a gaussian denoiser:
  Residual learning of deep cnn for image denoising. TIP, 2017

\bibitem{zhang2017learning}
Zhang, K., Zuo, W., Gu, S., Zhang, L.: Learning deep cnn denoiser prior for
  image restoration. In: CVPR, 2017

\bibitem{zhang2018ffdnet}
Zhang, K., Zuo, W., Zhang, L.: Ffdnet: Toward a fast and flexible solution for
  {CNN} based image denoising. TIP, 2018

\bibitem{zhangrdn2020}
Zhang, Y., Tian, Y., Kong, Y., Zhong, B., Fu, Y.: Residual dense network for
  image restoration. TPAMI, 2020

\bibitem{domainplus}
Zheng, B., Pan, X., Zhang, H., Zhou, X., Slabaugh, G., Yan, C., Yuan, S.:
  Domainplus: Cross transform domain learning towards efficient high dynamic
  range imaging. In: ACM MM, 2022

\bibitem{zheng2020image}
Zheng, B., Yuan, S., Slabaugh, G., Leonardis, A.: Image demoireing with
  learnable bandpass filters. In: CVPR, 2020

\bibitem{zheng2021learning}
Zheng, B., Yuan, S., Yan, C., Tian, X., Zhang, J., Sun, Y., Liu, L., Leonardis,
  A., Slabaugh, G.: Learning frequency domain priors for image demoireing.
  TPAMI, 2021

\bibitem{zhu2017joint}
Zhu, L., Fu, C.W., Lischinski, D., Heng, P.A.: Joint bi-layer optimization for
  single-image rain streak removal. In: ICCV, 2017

\bibitem{zhu2015fast}
Zhu, Q., Mai, J., Shao, L.: A fast single image haze removal algorithm using
  color attenuation prior. TIP, 2015

\bibitem{zhu1997prior}
Zhu, S.C., Mumford, D.: Prior learning and gibbs reaction-diffusion. TPAMI,
  1997

\bibitem{Zhu2021DeformableDD}
Zhu, X., Su, W., Lu, L., Li, B., Wang, X., Dai, J.: Deformable detr: Deformable
  transformers for end-to-end object detection. In: ICLR, 2021

\end{thebibliography}

\newpage
\appendix
\section{Appendix}

\subsection{Comparison with general image restoration methods}
We compare our TAPE-Net with two models, DnCNN~\cite{zhang2017beyond} and UNet~\cite{ronneberger2015u}, which are widely used in image restoration tasks. As shown in Table \ref{tab:expresults}, our model outperforms these two models on all datasets. Note that for a fair comparison, we set the compared models with a similar number of FLOPs.
The parameters of DnCNN, UNet, and our TAPE are 0.5M, 9.8M, and 1.3M, respectively. 
 
\begin{table}[h]
  \begin{center}
  \caption{Quantitative comparison for three models (in terms of PSNR (dB)).}
  \label{tab:expresults}
  \resizebox{12.0cm}{!}{
  \begin{tabular}{c|ccccc|cc|c}
    \toprule
      Dataset & Rain200L & Rain200H &Raindrop800&SIDD& TIP2018  &Snow100K  &ISTD &FLOPs \\
    \midrule 
    DnCNN~\cite{zhang2017beyond} &27.73&19.20&26.12&34.31&24.74& 23.77 &23.21 &14.4 G \\
    UNet~\cite{ronneberger2015u} &31.50&22.50&26.41&33.72&25.58& 23.48 & 25.64&18.2 G \\

    TAPE-Net (Ours) &\textbf{33.17} &\textbf{23.84} &\textbf{27.69} &\textbf{37.90} &\textbf{27.52} &\textbf{26.33} &\textbf{26.57}  &14.1 G\\

    \bottomrule
  \end{tabular}}
  \end{center}
\end{table}

\subsection{Comparison with task-specific methods}

As shown in Table~\ref{tab:denoise}, we firstly compare our methods with state-of-the-art denoising methods (DnCNN~\cite{zhang2017beyond}, FFDNet~\cite{zhang2018ffdnet}, RDN~\cite{zhangrdn2020}, and SADNet~\cite{chang2020spatial}). Note that the FLOPs number of our method (14.1G) is much smaller than that of RDN (46.6G) or SADNet (45.8G), when the input is a $64 \times 64$ RGB image.
We replace our backbone model (original transformer) with a specially designed transformer, Swin transformer~\cite{ZeLiu2021SwinTH} (termed as `TAPE-Net-swin-L' in Table~\ref{tab:denoise}), which outperforms all the SOTA methods with our pre-training strategy.

And we also compare with SOTA deraining methods in Table~\ref{tab:derain} and compare with SOTA demoireing methods in Table~\ref{tab:demoire}.
Our method outperforms all these task-specific SOTA methods.

\begin{table}[h]
  \begin{center}
  \caption{Quantitative comparison with the state-of-the-art denoising methods on SIDD.}
  \label{tab:denoise}
  \resizebox{12.0cm}{!}{
  \begin{tabular}{cccccccc }
    \toprule
       Method&DnCNN~\cite{zhang2017beyond}& FFDNet~\cite{zhang2018ffdnet} &RDN~\cite{zhangrdn2020} & SADNet~\cite{chang2020spatial}&TAPE-Net (Ours) & TAPE-Net-swin-L (Ours)\\

    \midrule 
      PSNR/SSIM & 34.31/0.892&33.26/0.890 &38.70/0.901 &38.41/0.900&37.90/0.896&38.76/0.901\\
      FLOPS (G)&14.4 &0.87&46.6& 45.8 &14.1&5.3\\
      
    \bottomrule
  \end{tabular}}
  \end{center}
\end{table}

\begin{table*}[h]
  \begin{center}
  \caption{Quantitative comparison with the state-of-the-art deraining methods on Rain200L and Rain200H. The best result are in \textbf{Bold}.}
  \label{tab:derain}
  \resizebox{12.4cm}{!}{
  \begin{tabular}{c|cccccccc }
    \toprule
       Dataset&Method&DDN~\cite{fu2017removing}& SPANet~\cite{wang2019spatial}&RESCAN~\cite{li2018recurrent} &PreNet~\cite{ren2019progressive} & BRN~\cite{ren2020single}&PCNet~\cite{2021pcnet}&TAPE-Net (Ours)\\
    \midrule 
     Rain200L& PSNR/SSIM & 28.35/0.878&30.92/0.930 &32.07/0.949&31.98/0.948 &32.40/0.953&32.62/0.954&\textbf{33.17}/\textbf{0.959}\\

     \midrule 
     Rain200H&PSNR/SSIM &20.98/0.705 &22.65/0.714 &23.04/0.729 &23.27/0.743 & 23.39/0.755&23.43/0.755&\textbf{23.84}/\textbf{0.759} \\

    \bottomrule
  \end{tabular}}
  \end{center}
\end{table*}

\begin{table}[h]
  \begin{center}
  \caption{Quantitative comparison with the state-of-the-art dermoireing methods on TIP2018. The best result are in \textbf{Bold}.}
  \label{tab:demoire}
  \resizebox{12.4cm}{!}{
  \begin{tabular}{cccccccc }
    \toprule
      Method & DMCNN~\cite{sun2018moire} & MopNet~\cite{he2019mop} & HRDN~\cite{yang2020high}&  FHDe2Net~\cite{hefhde2net}& WDNet~\cite{liu2020wavelet}& MBCNN~\cite{zheng2021learning} & TAPE-Net (ours)\\
    \midrule 
       PSNR/SSIM&25.82/0.806 & 26.20/0.861 &26.68/0.864&26.25/0.862 & 26.86/0.865& 27.37/0.866 &\textbf{27.52}/\textbf{0.866}\\
       
    \bottomrule
  \end{tabular}}
  \end{center}
\end{table}

\subsection{Additional visualization results.}

\subsubsection{Visualization some results of the prior queries, $Q$.}

\begin{figure}[h]
		\centering
		\includegraphics[width=0.80\textwidth]{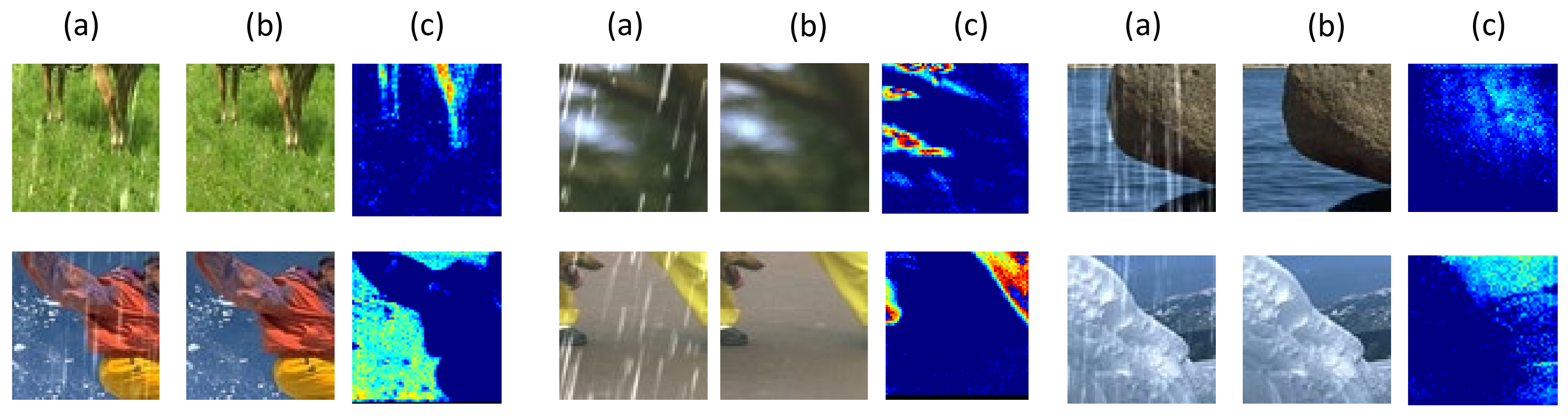}
		\caption{Visualization of some results of the prior queries, $Q$ on the Rain200L dataset. (a) Rain inputs. (b) Ground truth. (c) one of the predicted results of PLM.}
		\label{fig:visualize_q}
\end{figure}

We also visualize some other results of the prior queries ($Q$) of TAPE.
As shown in Fig. \ref{fig:visualize_q}, (a) and (b) are rain inputs and ground truth respectively; (c) are one of the predicted results of PLM.
We can see that with the help of pre-training, the PLM module can correlate the information of similar textures or patches from a long distance.
Thus, the transformer decoder of the backbone can utilize these long-distance similar areas/patches to restore the image.

\subsubsection{Visualization some results of learned parameters ($e$).}

\begin{figure}[h]
		\centering
		\includegraphics[width=0.80\textwidth]{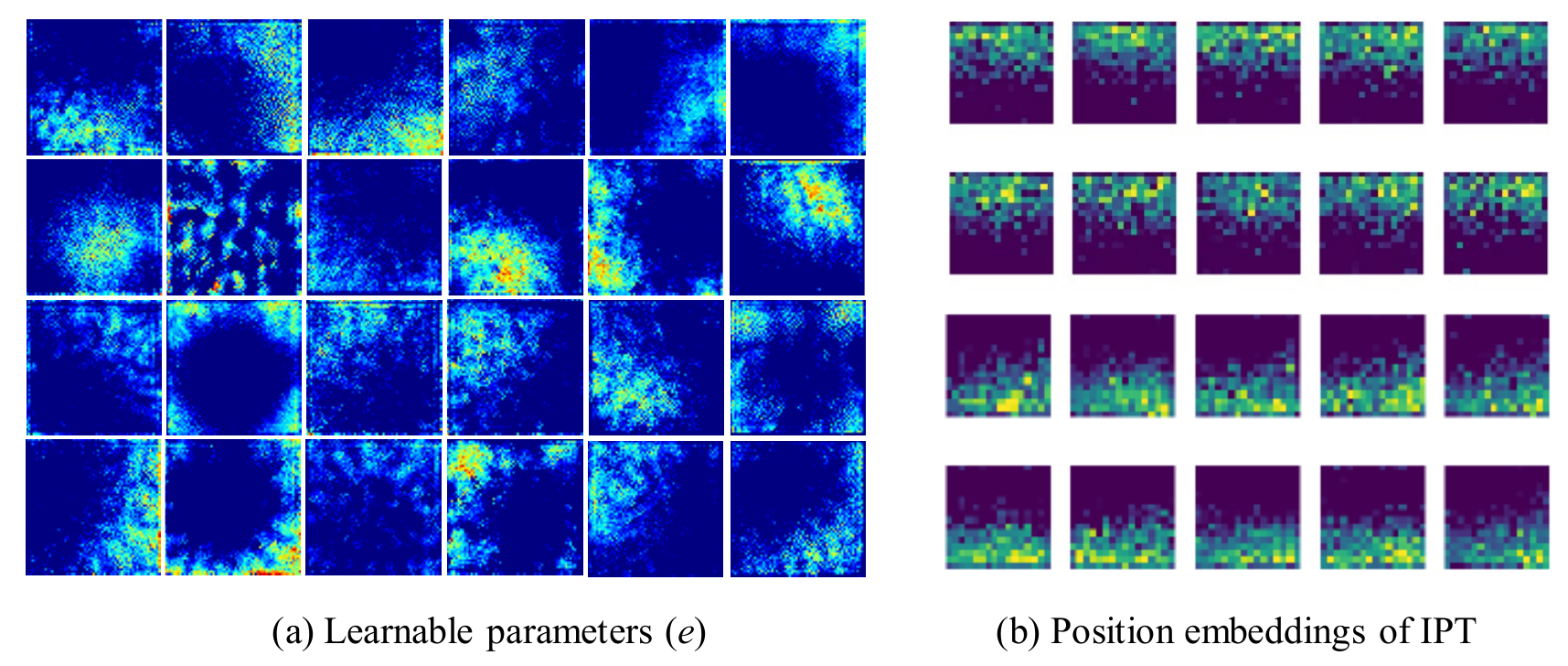}
		\caption{Visual comparison between our learnable parameters and the position embeddings of IPT. Our learnable parameters show more richer patterns.}
		\label{fig:visualize_e}
\end{figure}

We visualize the learned parameters ($e$) of TAPE. Fig. \ref{fig:visualize_e}
shows some visualization results of learned parameters ($e$) and the position embeddings of IPT (these results are copied from~\cite{chen2020pre}). 
We can find that our learned parameters ($e$) focus on other patches with farther distances than the position embeddings of IPT.
Besides, our learned feature maps show richer patterns. For example, some of the patches focus on the four corners of the image at the same time (the patch on the 3rd row and 2nd column). Some of the patches focus on the characters of oblique directions (the patch on the 1st row and 5th column).
These rich feature maps do not appear in the visualization results of IPT.

\subsection{Additional results on other transformer backbones.}
Our TAPE is a widely applicable method, where the backbone can be replaced by other transformer backbones.
We replace our backbone network with the swin transformer backbone~\cite{ZeLiu2021SwinTH} for experiments.
As shown in Table~\ref{tab:othernetwork}, `Baseline-swintrans' is the Swin transformer backbone without pre-training. `TAPE-Net-swintrans-S', `TAPE-Net-swintrans-M', and `TAPE-Net-swintrans-L' are our TAPE-Nets with the Swin transformer backbone and contain 1 Swin block, 3 Swin blocks and 5 Swin blocks, respectively. 
The results illustrate that our 3-stage pre-training and adding blocks can significantly boost the performance.

\begin{table}[h]
  \begin{center}
  \caption{Quantitative comparison for Baseline-swintrans and TAPE-Net-trans  (in terms of PSNR (dB)). The numbers in () of the 2nd line are the PSNR gain compared with `Baseline-swintrans'.}
  \label{tab:othernetwork}
  \resizebox{12.2cm}{!}{
  \begin{tabular}{c|cccccc}
    \toprule
                & Blocks numbers & Network parameters & Rain200L (dB) &SIDD (dB) &Raindrop800 (dB) \\

    \midrule 
    Baseline-swintrans& 1&0.19M&33.52&38.01&27.79\\
    TAPE-Net-swintrans-S&1&0.19M&34.07 (+0.55) &38.76 (+0.75) &28.31 (+0.52)  \\
    \midrule
    TAPE-Net-swintrans-M&3&0.61M&34.20 & 38.87&28.97\\
    TAPE-Net-swintrans-L&5&0.97M&34.46  & 38.98 &29.15\\

    \bottomrule
  \end{tabular}}
  \end{center}
\end{table}

\subsection{Ablation study about optimization in task-specific fine-tuning.}
In Sec. 3.2.1, there are two ways to optimize the networks in the task-specific fine-tuning stage,
namely: 1) The backbone $\phi$ is fine-tuned by loss between pseudo GT and GT
firstly, and then fixed when fine-tuning other networks (denoted as Step by Step finetuning in Table~\ref{tab:opt_abstudy}); 2) All components are
fine-tuned simultaneously (denoted as Joint finetuning in Table~\ref{tab:opt_abstudy}). As shown in Table~\ref{tab:opt_abstudy}, the performance of the two methods is equivalent. How to choose different optimization methods for different tasks is future work.

\begin{table}[t]
  
\begin{center}
  \caption{Comparison between Step by step fine-tuning and Joint fine-tuning.}
  \label{tab:opt_abstudy}
  \resizebox{8.8cm}{!}{
  \begin{tabular}{cccc}
    \toprule
      TAPE-Swin & Rain200L & Rain200H &  Raindrop800-TestB \\
    \midrule 
     Step by step fine-tuning & 35.10 & 26.18 & 26.41\\

      Joint fine-tuning& 35.06 & 26.05 & 26.49 \\
    \bottomrule
  \end{tabular}}
  \end{center}
\end{table}

\subsection{Ablation Study of pixel-wise contrastive loss.}
We remove the pixel-wise contrastive loss in the task-agnostic pre-training. And the PSNR/SSIM decrease by 0.12dB/0.001 on Rain200L without the proposed pixel-wise contrastive loss.

\subsection{Ablation Study about Transformer or CNN.}

We replace the transformer encoder and decoder with the ResNet encoder and decoder \cite{resunet} respectively with the same model size. The feature map outputed from the encoder and the feature map outputed from the PLM are concatenated and served as the input of the ResNet decoder. 
We do the ablation study on the Rain200L dataset with the same setting as Sec. 4.5 of the main paper. 
The PSNR/SSIM drops 1.03dB/0.006 compared with the baseline with the transformer encoder and decoder.
The result shows that using the transformer is better when fusing the information of the output of PLM and the encoder.
We also made a comparison between pre-trained CNN and no-pre-trained CNN. Compared with transformer, CNN's performance improvement is much smaller.

\subsection{Visual comparison with other SOTA methods on desnowing and shadow removal}
\begin{figure}[h]
		\centering
		\includegraphics[width=0.80\textwidth]{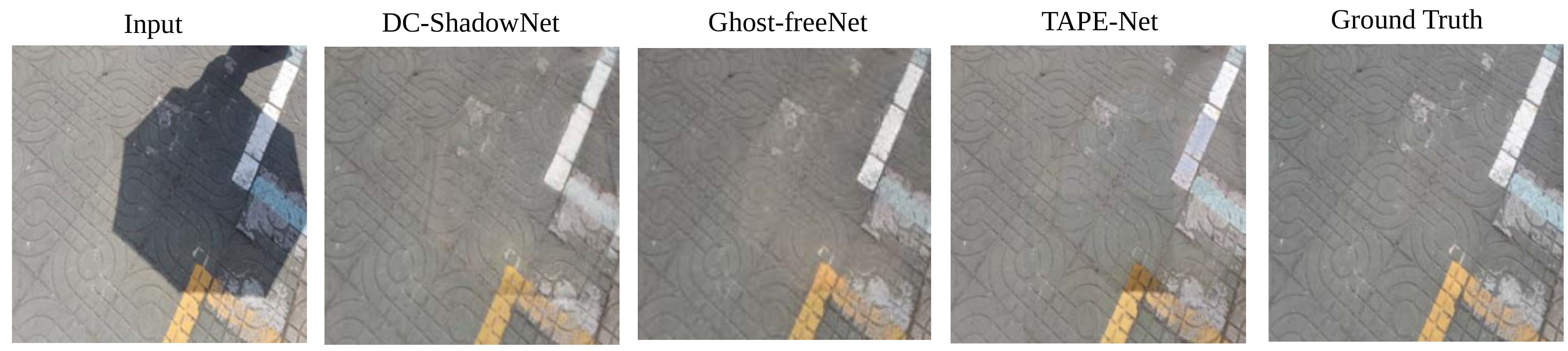}
		\caption{Visual shadow removal comparison among ours and two other methods (DC-ShadowNet~\cite{jin2021dc} and Ghost-freeNet~\cite{cun2019ghostfree}).}
		\label{fig:deshadow}
\end{figure}

\begin{figure}[h]
		\centering
		\includegraphics[width=0.80\textwidth]{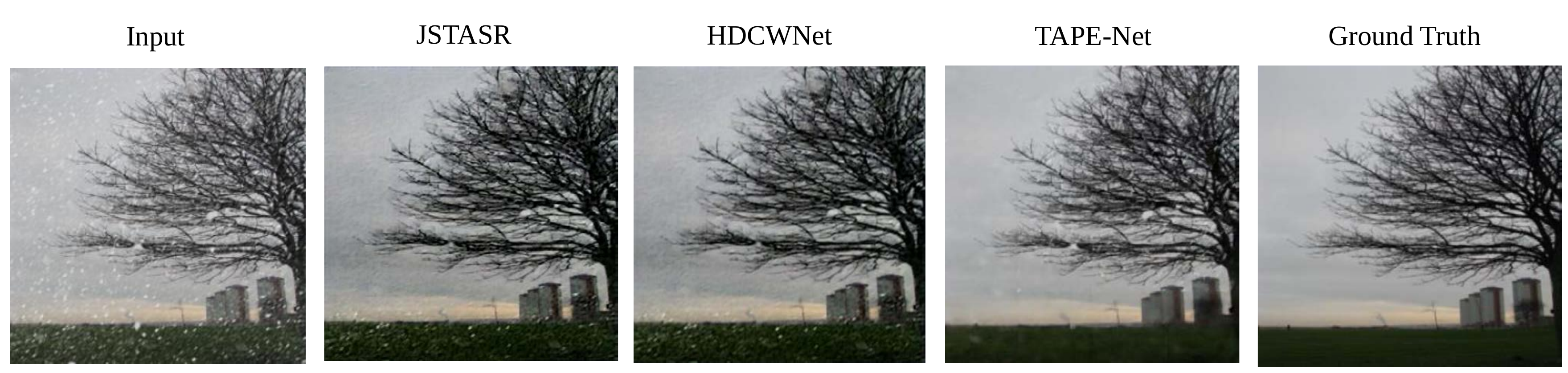}
		\caption{Visual desnowing comparison among ours and two other methods (JSTASR~\cite{chen2020jstasr} and HDCWNet~\cite{chen2021all}).}		\label{fig:desnow}
\end{figure}

We compare our method with several state-of-the-art desnowing and shadow removal methods. Fig.~\ref{fig:deshadow} and Fig.~\ref{fig:desnow} show our method can remove the shadow or snow. Please note that these compared methods use the snow/shadow masks for training, while our method only uses snow/snow-free or shadow/shadow-free image pairs.

\end{document}